# A modified fuzzy C means algorithm for shading correction in craniofacial CBCT images

Awais Ashfaq[1] and Jonas Adler[2]

[1] KTH Royal Institute of Technology and Halmstad University
[2] KTH Royal Institute of Technology and Elekta Instrument AB

**Abstract.** CBCT images suffer from acute shading artifacts primarily due to scatter. Numerous image-domain correction algorithms have been proposed in the literature that use patient-specific planning CT images to estimate shading contributions in CBCT images. However, in the context of radiosurgery applications such as gamma knife, planning images are often acquired through MRI which impedes the use of polynomial fitting approaches for shading correction. We present a new shading correction approach that is independent of planning CT images. Our algorithm is based on the assumption that true CBCT images follow a uniform volumetric intensity distribution per material, and scatter perturbs this uniform texture by contributing cupping and shading artifacts in the image domain. The framework is a combination of fuzzy C-means coupled with a neighborhood regularization term and Otsu's method. Experimental results on artificially simulated craniofacial CBCT images are provided to demonstrate the effectiveness of our algorithm. Spatial non-uniformity is reduced from 16% to 7% in soft tissue and from 44% to 8% in bone regions. With shading-correction, thresholding based segmentation accuracy for bone pixels is improved from 85% to 91% when compared to thresholding without shading-correction. The proposed algorithm is thus practical and qualifies as a *plug and play* extension into any CBCT reconstruction software for shading correction.

## 1 Introduction

The development of compact 2D detector arrays coupled with a modified filtered backprojection algorithm [1] that supports volumetric reconstruction from 2D projections acquired in a circular trajectory led to the introduction of *Cone Beam Computed Tomography* (CBCT)[2]. CBCT expanded the role of x-ray imaging from medical diagnosis to image guidance in operative and surgical procedures. C-arm mounted CBCT is another advancement in radiology which allows CBCT system to be physically attached to therapeutic devices such as proton therapy, linear accelerators and gamma knife [3,4].

Since the introduction of CBCT, there has been significant research to address scatter noise. These can be broadly classified into hardware and software approaches. Software approaches can be further divided into reconstruction and post-reconstruction techniques. Details on existing hardware and software methods for shading correction can be found in section A of our previous work [5]. In this work we focus on post-reconstruction techniques that eliminate shading artifacts from previously reconstructed images. They include image transformation algorithms coupled with prior information - usually a CT image - to minimize the difference between the reconstructed CBCT image and CT image of the same patient. [6,7,8,9]. The common idea is to build a mathematical fitting model that minimizes a distance metric between the CT and CBCT pixel values. In radiotherapy applications, the CT information is gathered from planning images captured by conventional fan beam CT. However, it is desirable to replace CT planning images with MR images due to higher soft tissue contrast and no radiation exposure [10]. In the context of radiosurgery applications such as gamma knife, planning images are rarely acquired through CT scans, rather, MR images are used. This impedes the general use of polynomial fitting approaches for shading correction in CBCT images.

Recently P. Wei et al. proposed an algorithm for CBCT shading correction without patient-specific prior information [11]. However, the algorithm requires an initially hard segmented CT template image to attain general anatomical information. The ideal template is then subtracted from the input CBCT image to achieve a residual image carrying different error sources. The residual image is then forward projected and low-frequency shading artifacts are filtered in the projection domain. The steps are iteratively repeated until the variance of residual image is minimized.

We present a new approach for shading correction in CBCT images that is independent of prior information and operates in the image domain. The framework is a combination of fuzzy C-means (FCM) [12] coupled with a neighborhood regularization term and Otsu's segmentation method [13,14]. It is based on the assumption that true CBCT images follow a uniform volumetric intensity distribution per material, and scatter perturbs this uniform texture by contributing cupping and shading artifacts in the image domain. The neighborhood regularization term within the FCM objective function biases the solution to piece-wise smooth clusters. Otsu's method initializes the cluster centers to ensure similar solution every time the function is run. Accurate hard segmentation of the final image is of interest for applications such as dose calculation [15], or estimating scatter convolution kernels in analytical reconstruction [16] etc. To achieve this, we propose a Bayesian framework [17] that may be used with prior spatial probabilistic maps for the final segmentation to be in consistent with expected craniofacial anatomy.

The basic framework is inspired from the analogy between *shading* and *bias-field effect* - which is a low frequency noise signal - observed in MR images. Several variants of FCM have been proposed in the literature for estimating bias-field in MR images [18,19,20,21]. We extend the application of those frameworks on craniofacial CBCT images. To the best of our knowledge, no similar approaches have been applied for CBCT shading correction. Henceforth, we shall use the term *bias-field* and *shading* interchangeably to elaborate our approach. The general framework for estimating bias-field is independent of prior information and does not involve complex and computationally expensive physical or mathematical models. It is a post-reconstruction algorithm operating in image domain.

## 2 Methods

The overall framework is illustrated in Fig. 1. The algorithm performs bias correction directly on volumetric CBCT images.

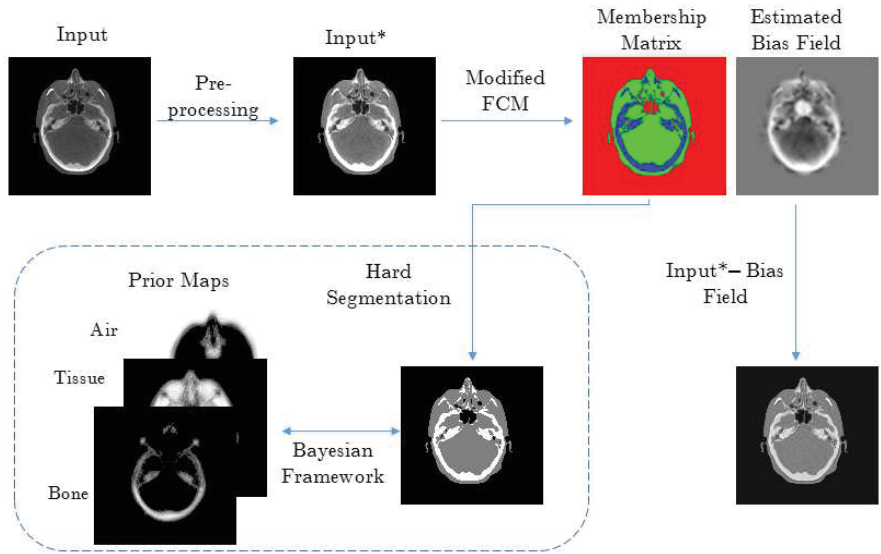

Fig. 1: A general illustration of the proposed bias correction framework. Steps within the dotted rectangle are not part of bias-field estimation and will be discussed later in the paper.

In the pre-processing step, CBCT volume is first converted into a binary image. The largest contiguous volume is separated from the background using morphological operations and the background pixels are set to zero. This step eliminates the background noise and improves FCM performance. Outliers are discarded by selecting intensity values between 5% and 85% of intensity spectrum. The resulting image is normalized to the range [0 , 1].

### 2.1 Bias-field estimation

The observed CBCT image is modelled as a sum of true signal and a spatially slow varying bias-field.

$$y_n = x_n + b_n \quad \forall\, n \in \{1, 2, 3, ...., N\} \tag{1}$$

where
$N$ is the number of pixels in the image
$y_n$ is the observed intensity at $n^{th}$ pixel
$x_n$ is the true intensity at $n^{th}$ pixel
$b_n$ is the bias intensity at $n^{th}$ pixel.

The FCM objective function is given by

$$J = \sum_{i=1}^{I} \sum_{n=1}^{N} \mu_{in}^{m} \, \|x_n - c_i\|^2 \tag{2}$$

where
$I$ is the number of clusters
$c_i$ is the center of the $i^{th}$ cluster
$\mu_{in}$ is the degree of membership of $x_n$ in the $i^{th}$ cluster and $\mu_{in} \in [0,\,1]$.
$m$ is the level of fuzziness. $m \in (1,\,\infty)$. $m = 1$ would imply a strict partitioning between clusters. Higher values of $m$ imply smaller membership values and hence fuzzier clusters. For a given data point, sum of the membership values for every cluster is normalized to 1.

$$\sum_{i=1}^{I} \mu_{in} = 1 \quad \forall\, n \tag{3}$$

The modified FCM objective function as proposed by Ahmed et al [18] is given by

$$J^* = J + \frac{\alpha}{|M|} \sum_{i=1}^{I} \sum_{n=1}^{N} \mu_{in}^{m} \left( \sum_{r \in M_n} \|x_r - c_i\|^2 \right) \tag{4}$$

where
$M_n$ is the set of neighbours around $x_n$
$|M|$ is a constant denoting the cardinality of $M_n$
$\alpha \in \mathbb{R}$ controls the neighborhood influence.

In CBCT acquisition, intensity values correspond to attenuation coefficients of the material being scanned which in turn roughly correspond to the material's density. It is thus important to have a piece-wise smooth solution while keeping the true intensity values $x_n$ close to the observed intensity values $y_n$ (see Eq. 1). In order to preserve this information, we assume that the bias-field has zero mean.

$$\sum_{n=1}^{N} b_n = 0 \tag{5}$$

Substituting Eq. 1 in Eq. 4

$$J^* = \sum_{i=1}^{I} \sum_{n=1}^{N} \mu_{in}^m \, D_{in} + \frac{\alpha}{|M|} \sum_{i=1}^{I} \sum_{n=1}^{N} \mu_{in}^m \, R_{in} \qquad (6)$$

where
$D_{in} = \|y_n - b_n - c_i\|^2$ and $R_{in} = \sum_{r \in M_n} D_{ir}$
The task of minimizing $J^*$ in Eq. 6 with constraints given by Eq. 3 and 5 is a nonlinear optimization problem. Mathematically,

$$\underset{\mu,c,b}{\text{minimize}} \quad J^*$$
$$\text{subject to} \sum_{i=1}^{I} \mu_{in} = 1 \quad \forall\, n$$
$$\sum_{n=1}^{N} b_n = 0$$

Ahmed et al. [18] suggested an algorithm to minimize a similar optimization problem using Lagrange multipliers. We follow similar steps and define the lagrangian as

$$\mathcal{L} = J^* + \sum_{n=1}^{N} \gamma_n \left(1 - \sum_{i=1}^{I} \mu_{in}\right) + \lambda \sum_{n=1}^{N} b_n \qquad (7)$$

Taking the derivative of Eq. 7 with respect to $\mu_{in}, c_i, b_n$[1].,$\gamma_n$ and $\lambda$. For an optimal solution, all of these derivatives would be simultaneously zero. However, finding such a point is non-trivial and we resort to coordinate descent to find an optimal solution [22].

$$\frac{\partial \mathcal{L}}{\partial \mu_{in}} = m\mu_{in}^{m-1} D_{in} + \frac{\alpha m}{|M|} \mu_{in}^{m-1} R_{in} - \gamma_n$$

$$\frac{\partial \mathcal{L}}{\partial c_i} = 2 \sum_{n=1}^{N} \mu_{in}^m \, (b_n + c_i - y_n) + \frac{2\alpha}{|M|} \sum_{n=1}^{N} \mu_{in}^m \left(\sum_{r \in M_n} (b_r + c_i - y_r)\right)$$

$$\frac{\partial \mathcal{L}}{\partial b_n} = 2 \sum_{i=1}^{I} \mu_{in}^m \, (b_n + c_i - y_n) + \frac{2\alpha}{|M|} \sum_{i=1}^{I} \sum_{r \text{ s.t. } n \in M_r} \mu_{ir}^m \, (b_n + c_i - y_n) + \lambda$$

$$\frac{\partial \mathcal{L}}{\partial \gamma_n} = 1 - \sum_{i=1}^{I} \mu_{in}$$

$$\frac{\partial \mathcal{L}}{\partial \lambda} = \sum_{n=1}^{N} b_n$$

---

[1] Please note that $\frac{\partial \mathcal{L}}{\partial b_n}$ differs from Ahmed et al. [18] because the latter neglects contributions from neighboring pixels of $x_n$.

Update equation for membership matrix $\mu$ is given by

$$\mu_{in} = \left( \frac{\gamma_n}{m \left( D_{in} + \frac{\alpha}{|M|} R_{in} \right)} \right)^{\frac{1}{m-1}} \quad (8)$$

and $\gamma_n$ is evaluated using constraint Eq. 3.

$$\gamma_n = \frac{m}{\left( \sum_{j=1}^{I} \left( \frac{1}{D_{jn} + \frac{\alpha}{|M|} R_{jn}} \right)^{\frac{1}{m-1}} \right)^{m-1}} \quad (9)$$

Update equation for cluster center $c$ is given by

$$c_i = \frac{\sum_{n=1}^{N} \mu_{in}^m \left( (y_n - b_n) + \frac{\alpha}{|M|} \sum_{r \in M_n} (y_r - b_r) \right)}{(1 + \alpha) \sum_{n=1}^{N} \mu_{in}^m} \quad (10)$$

Update equation for bias-field $b$ is given by

$$b_n = y_n - \frac{\sum_{i=1}^{I} c_i \beta_{in} + \lambda/2}{\sum_{i=1}^{I} \beta_{in}} \quad (11)$$

where

$$\beta_{in} = \mu_{in}^m + \frac{\alpha}{|M|} \sum_{r \text{ s.t. } n \in M_r} \mu_{ir}^m$$

and $\lambda$ is evaluated using constraint Eq. 5.

$$\lambda = 2 \left( \sum_{n=1}^{N} \frac{1}{\sum_{i=1}^{I} \beta_{in}} \right)^{-1} \sum_{n=1}^{N} \left( y_n - \frac{\sum_{i=1}^{I} c_i \beta_{in}}{\sum_{i=1}^{I} \beta_{in}} \right) \quad (12)$$

The algorithm is summarized as below[2]

1. Determine initial cluster centers $c_i \ \forall \ i$ using Otsu's method.
2. Randomly initialize $b_n \ \forall \ n$ to small values.
3. Update partition matrix $\mu_{in}$ using Eq. 8.
4. Update cluster centers $c_i$ using Eq. 10.
5. Update bias term $b_n$ using Eq. 11.
6. Repeat steps $3 - 5$ till termination criteria as follows.

$$\|c_{new} - c_{old}\| < \epsilon$$

where
$c$ is an array of cluster centers
$\epsilon$ is a small user defined constant.

The estimated bias-field is then passed through a 3D Gaussian kernel to smooth out the correction values. It is then subtracted from the input image to obtain the estimated true values.

---

[2] For source codes, visit `https://github.com/adler-j/mfcm_article`.

## 2.2 Data Generation

CBCT images were acquired in the following order. More details can be found in section 2.1.1-I of [5].

1. Craniofacial real CT data was provided by Elekta, Stockholm - Data A.
2. Data A was hard segmented into bone, tissue and air regions via multilevel-thresholding [14] - Data B.
3. CBCT projections were then acquired using Elekta's proprietary algorithm involving density calculations from CT Data A, segmented CT Data B and CBCT energy spectrum of the Elekta's Leksell Gamma Knife Icon. This model uses Monte-Carlo simulations to account for physical processes involved in the acquisition process such as detector response, scatter and beam hardening.
4. Finally, CBCT projections were reconstructed using conjugate gradient least square technique in ODL[3].

## 2.3 Evaluation

The aforementioned algorithm was tested on artificially simulated 2D Forbild phantom image from ODL and 3D craniofacial CBCT images. A major advantage of using simulated images is the availability of true knowledge of tissue type. This helps in an accurate quantitative evaluation of the algorithm. Material indices from Data B were used to extract respective material pixels from original and bias-corrected CBCT images. Standard deviation among pixels of similar material type was calculated - $std(I_{tissue})$. A standard deviation of 0 indicates a totally flat tissue region. Smaller values of $std(I_{tissue})$ are preferred.

We also evaluated the effectiveness of the proposed algorithm through segmentation results based on multilevel-thresholding for original and bias-corrected CBCT images. Histograms based Otsu's method was used to determine thresholds [13,14]. It must be noted that shading artifacts disrupt image-domain histograms leading to inaccurate segmentation thresholds. Sensitivity and specificity values were calculated using Eq. 13 and 14. The standard two sample T-test [24] was used to determine if mean segmentation error for original and bias-corrected images was statistically different from one another.

$$\text{Sensitivity}_{tissue} = \frac{\text{No. of true positives}}{\text{No. of true positives} + \text{No. of false negatives}} \quad (13)$$

$$\text{Specificity}_{tissue} = \frac{\text{No. of true negatives}}{\text{No. of true negatives} + \text{No. of false positives}} \quad (14)$$

---

[3] Available at https://github.com/odlgroup/odl

## 3   Results

The aforementioned algorithm was implemented on Matlab R2016a running on Intel Core i7-4790K CPU. Figure 2 demonstrates its ability in reducing cupping artifact visible around central pixels in Forbild phantom image.

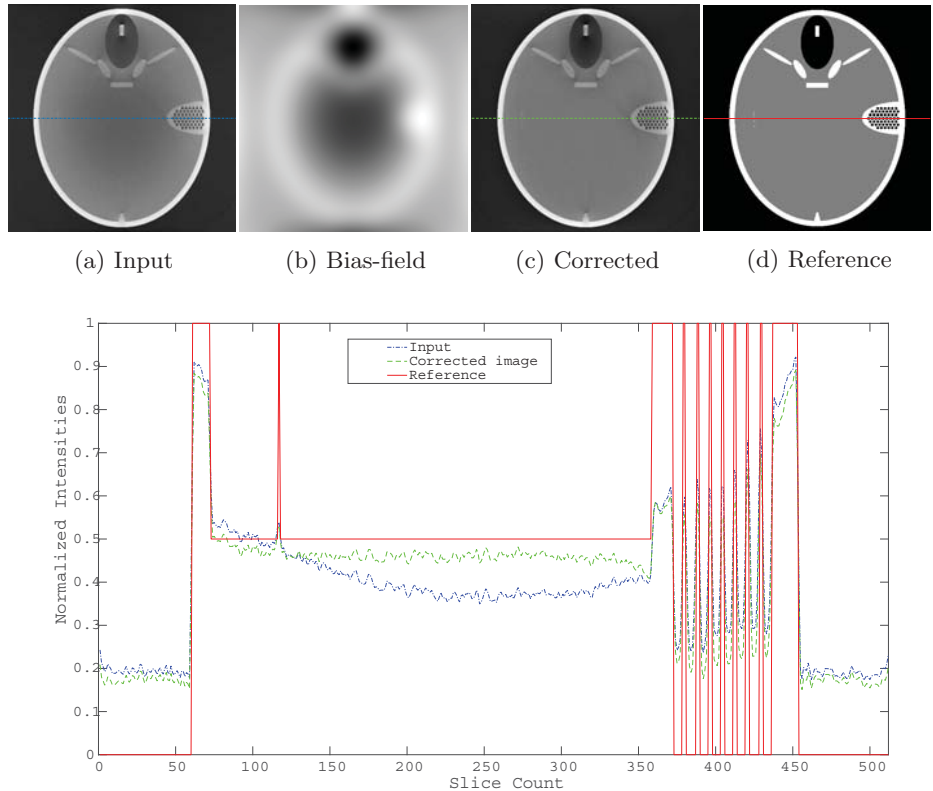

Fig. 2: Horizontal line profile illustrating shading correction in Forbild image.

### 3.1   Application on craniofacial CBCT images

To ensure robustness and result fidelity, CBCT images were divided into training and test images. Parameter tuning was performed on 5 training images and the optimized model - Table 1 - was then applied on 20 test images. For computational reasons, we chose a neighborhood that lies in the same slice as the pixel of interest.

The non-uniformity factor depending on standard deviation was reduced from 16% to 7% in tissue regions and from 44% to 8% in bone regions - see Figure 3.

Table 1: Suggested parameter values of the proposed algorithm

| $I$ | $m$ | $\alpha$ | $|M|$ | $\epsilon$ |
|---|---|---|---|---|
| 3 | 2 | 1 | $(3 \times 3) - 1$ | $10^{-5}$ |

This is visualized in Figure 4 which provides a comparison between original and bias-corrected CBCT images.

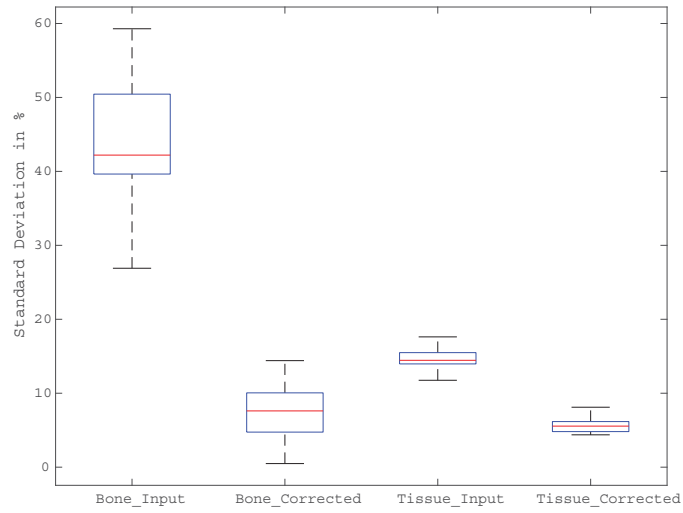

Fig. 3: Box plot showing statistical distribution of standard deviations among different tissue regions before and after bias correction evaluated over 20 test images. For each box, the red mark indicates the median, and the bottom and top edges of the box indicate the $25^{th}$ and $75^{th}$ percentiles, respectively. Whiskers extend to extreme data points that are not considered as outliers.

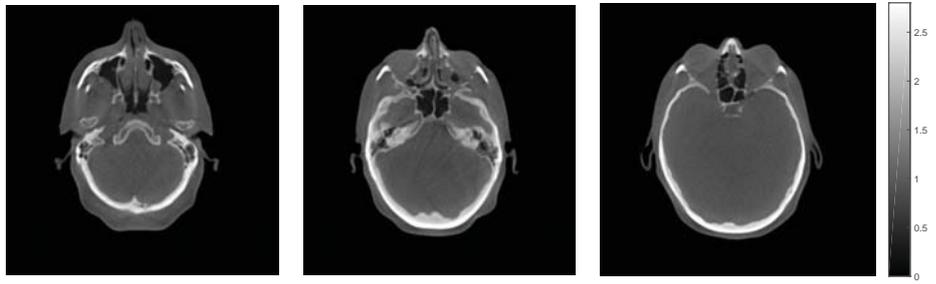

A. Input slices

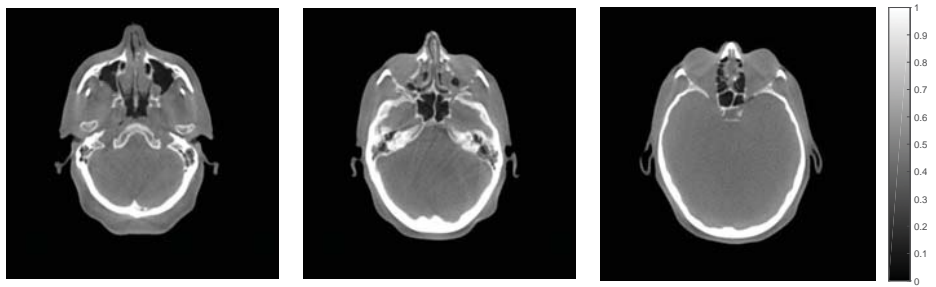

B. Input* slices (after pre-processing)

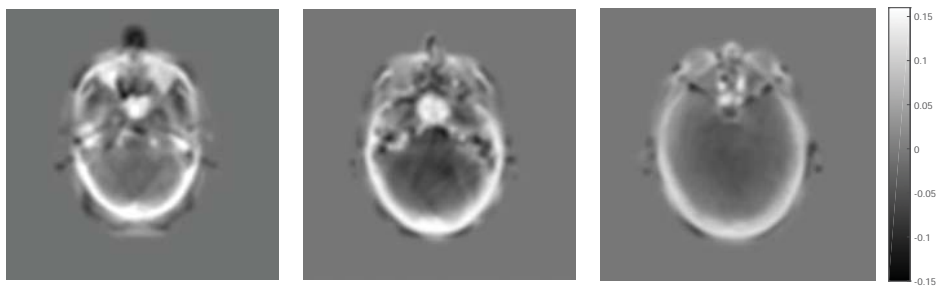

C. Estimated bias-field

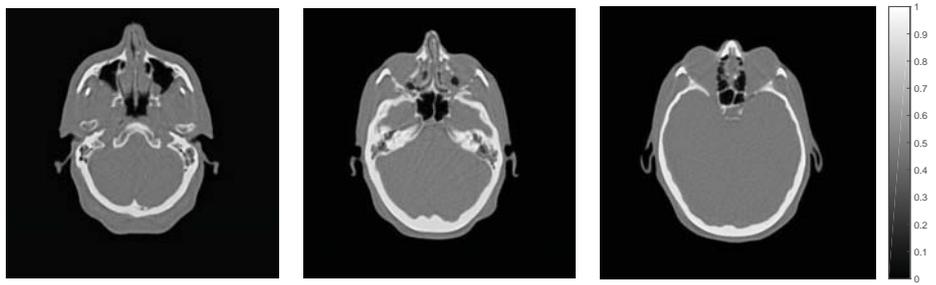

D. Corrected slices

Fig. 4: Visual inspection of different craniofacial slices. Published with permission from Elekta Instrument AB.

It is important to note that biasing the solution to piece-wise smooth clusters using the regularizing neighborhood term results in slight deterioration of edges. Risk of losing sharp edges within image is increased with a larger window size. The number of FCM clusters was initialized to 3 (bone, tissue and air) in the current implementation - Table 1. An implication of this configuration is that muscle-fat or soft-hard bone boundaries are smoothed out - Figure 5. Further efforts to optimize the model with more clusters are required to have distinctive boundaries between tissues with minimal density differences. It is also important to note that for computational reasons, no smoothing constraint is imposed on bias-field $b$ within the proposed iterative algorithm. Rather a 3D Gaussian kernel is used to smooth the bias correction values. This low-pass filtering also deteriorates sharp edges in the image. Contrary to the existing bias constraint in Eq. 5, we also investigated the alorithm by imposing smoothness constraints on $b$ such as penalizing the *square of bias* and *square of the gradient of bias* - Tichonov regularization [23] - which forces the $b$'s to be *low* and *slow varying* respectively [4].

However the results were comparative in both cases except for the execution time where using the gradient constraint required significantly more iterations compared to the former proposed algorithm. It is also worth adding that in the context of radiosurgery, CBCT is often not meant for diagnostic studies, rather it is used to determine patient position prior to surgical sessions. Hence a high soft tissue contrast is not a strict requirement on CBCT images.

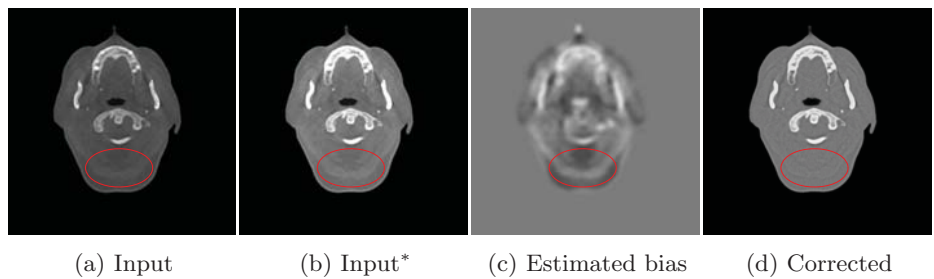

(a) Input  (b) Input*  (c) Estimated bias  (d) Corrected

Fig. 5: Smoothing effect marked in red across fat-tissue boundary. Image grayscale theme is same as in figure 4. Published with permission from Elekta Instrument AB.

Table 2 demonstrates multilevel-thresholding based segmentation performance on CBCT images with and without bias correction denoted as *BC* and *WBC* respectively. A noticeable improvement was visualized in the sensitivity value of bone pixels from 85.3% to 91.7%. This is seen as loss in contrast around central bone regions due to shading that result in missing bone pixels segmented through thresholding. This is also visualized in Figure 4 which provides a comparison between original and bias-corrected CBCT images. Following the decrease in mean error in Table 2, a statistical T-test was performed at 5% significance level. Based on a p-value < 0.001, it can be generalized that in 95% of cases, bias correction followed thresholding shall outperform simple thresholding operation on CBCT images.

---
[4] For mathematical derivations, visit https://github.com/adler-j/mfcm_article/blob/master/python/MFCM_continuum.pdf

Table 2: Evaluation metric averaged over 20 test images.

| Multilevel thresholding | Evaluation metric | | | | | | | | |
|---|---|---|---|---|---|---|---|---|---|
| | Sensitivity % | | | Specificity % | | | Mean error% | Standard deviation% | Time s |
| | Bone | Tissue | Air | Bone | Tissue | Air | | | |
| WBC | **85.3** | 99.3 | 99.7 | 99.9 | 98.8 | 99.6 | 1.1 | 0.18 | 0.9 |
| BC | **91.7** | 98.5 | 99.8 | 99.8 | 98.3 | 99.4 | 0.8 | 0.12 | 64 |

In a sense, the improvement in mean error ($\approx 0.3\%$) seems fairly low compared to execution time. However, it must be noted that 0.3% misclassification in volumetric images of size ($512 \times 512 \times 154$) imply approximately 120,000 misclassified pixels. Figure 6 demonstrates this effect by counting bone pixels segmented per slice using multilevel-thresholding with and without bias correction and gives a comparison against the ground truth. Significant loss in bone information is observed in central slices via thresholding without bias correction, which obscures crucial structural information.

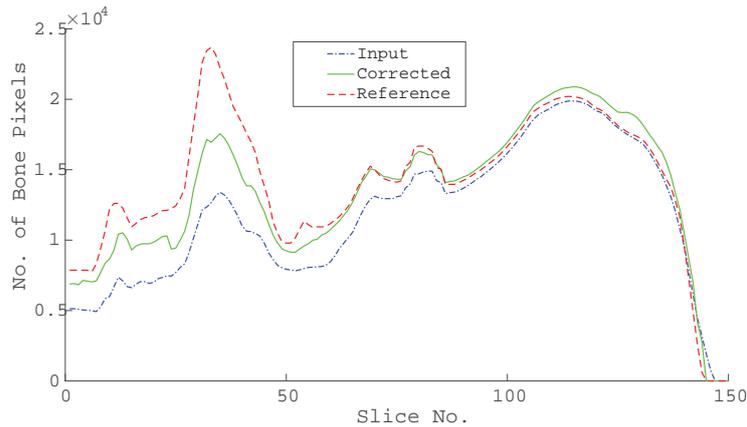

Fig. 6: Bone information per craniofacial CBCT slice in transverse plane. The highest slice No. corresponds to top of the head.

A next step towards improving segmentation performance will be to build probabilistic maps of craniofacial anatomy - Figure 1. It depends on the availability and quality of craniofacial CBCT images in database. Details on generating probabilistic maps can be found in section 2.2.1 of [5]. The membership matrix $\mu$ in Eq. 8 was normalized in the range $[0, 1]$. These membership values can thus be treated as probabilities of a given pixel belonging to a specific cluster. Given this likelihood information and prior probabilities from maps, a Bayesian framework [17] can be set up for the segmentation to be in consistent with expected craniofacial anatomy. More details on this will follow in near future.

The execution time of the algorithm is also essential to consider. Current Matlab implementation requires $\approx 64$ seconds to estimate bias-field in a CBCT image of size $512 \times 512 \times 154$. Also it operates on slices and does not take into account 3D connectivity of pixels in the image which may improve bias estimation, yet at higher computational costs. Further efforts on GPU implementation of the algorithm are needed to enable its use as a *shading-correction* extension in real-time CBCT image reconstruction.

## 4 Conclusion

Following the analogy between *bias-field effect* in MRI and *shading* in CBCT; we postulate that past research on bias-field estimation can be used for CBCT shading correction. An image-domain based low-frequency shading correction method for CBCT images has been proposed. Being distinctive from other approaches, this method improves spatial uniformity independent of prior information which qualifies it as a *plug and play* extension to any CBCT reconstruction software. In addition, the improvement in pixel-wise segmentation accuracy attained after bias correction, qualifies it as a potential candidate to be used within several state-of-the-art iterative denoising algorithms [25,26] which require an accurate segmentation of the volume into different materials.

### Conflict of interest and compliance with ethical requirements

The authors confirm that there are no conflicts of interest in relation to the submitted manuscript. All CBCT images in Fig 1, 4 and 5 are published with permission from Elekta Instrument, Stockholm. CBCT data provided by Elekta was fully anonymized.